\documentclass[11pt]{article}

\usepackage[letterpaper, margin=1in]{geometry}

\usepackage{times}
\usepackage{microtype}

\usepackage{amsmath}
\usepackage{amsfonts}
\usepackage{amssymb}
\usepackage{nicefrac}

\usepackage{graphicx}
\usepackage{xcolor}
\usepackage[colorlinks=true,linkcolor=blue,citecolor=blue,urlcolor=blue]{hyperref}
\usepackage{url}

\usepackage{booktabs}
\usepackage{array}
\usepackage{colortbl}
\usepackage{siunitx}
\definecolor{rowAvg}{HTML}{EAF2FB}   
\definecolor{rowImp}{HTML}{F1F1F1}   
\definecolor{grpA}{HTML}{F5F5F5}     
\definecolor{grpB}{HTML}{FFF6E5}     
\definecolor{grpC}{HTML}{FDECEC}     
\definecolor{grpD}{HTML}{E8F3EA}     
\definecolor{topAUC}{HTML}{1F4E9D}   
\newcolumntype{Y}{S[table-format=1.4]}
\newcolumntype{Z}{S[table-format=+1.2, table-number-alignment=center]}

\usepackage[round,authoryear]{natbib}
\usepackage{titlesec}
\titlespacing*{\section}{0pt}{1.2em}{0.6em}
\titlespacing*{\subsection}{0pt}{1em}{0.4em}



\title{Lightweight Complementary-Cue Fusion for Robust Video Face Forgery Detection}
\author{%
  Sunghwan Baek\thanks{\texttt{sunghwab@alumni.cmu.edu}} \\
  Carnegie Mellon University
  \and
  Tariq Anwaar\thanks{\texttt{tjaheerh@alumni.cmu.edu}} \\ 
  Carnegie Mellon University
  \and
  Karanveer Singh\thanks{\texttt{karanves@alumni.cmu.edu}} \\ 
  Carnegie Mellon University
  \and
  Rita Singh\thanks{\texttt{rsingh@andrew.cmu.edu}} \\
  Carnegie Mellon University
}

\date{}

\begin{document}
\maketitle

\begin{abstract}
Current face video forgery detectors use wide or dual-stream backbones. We show that a single, lightweight fusion of two handcrafted cues can achieve higher accuracy with a much smaller model. Based on the Xception baseline model (21.9 million parameters), we build two detectors: LFWS, which adds a 1x1 convolution to combine a low-frequency Wavelet-Denoised Feature (WDF) with a phase-spectrum channel derived from Spatial-Phase Shallow Learning (SPSL), and LFWL, which merges WDF with Local Binary Patterns (LBP) in the same way. This extra module adds only 292 parameters, keeping the total at 21.9 million, smaller than F3Net (22.5 million) and less than half the size of SRM (55.3 million). Even with this minimal overhead, the fused models increase the average area under the curve (AUC) from 74.8\% to 78.6\% on FaceForensics++ and from 70.5\% to 74.9\% on DFDC-Preview, gains of 3.8\% and 4.4\% over the Xception baseline. They also consistently outperform F3Net, SRM, and SPSL in eight public benchmarks, without extra data or test-time augmentation. These results show that carefully paired, handcrafted features, combined through the lightweight fusion block, can provide competitive robustness at a significantly lower cost than comparable frequency-based detectors. Our findings suggest a need to reevaluate scale-driven design choices in face video forgery detection.
\end{abstract}

\section{Introduction}

Powerful generative models, such as GANs and diffusion models, have made it easy to create convincing deepfakes; however, distinguishing them from real images remains a significant challenge \citep{schwarz2021frequency,dzanic2020fourier}. Many recent detectors look for artifacts in a single, narrow representation: Local Binary Patterns (LBP) find local micro-texture anomalies \citep{ojala2002multiresolution}, Fourier-based methods reveal spectral magnitude artifacts \citep{dzanic2020fourier}, and Spatial-Phase Shallow Learning (SPSL) targets up-sampling artifacts in the phase spectrum \citep{liu2021spsl}. Wavelet-based techniques, in contrast, show that \emph{low-frequency} structural cues from the wavelet approximation band are also useful \citep{wolter2021waveletpackets}. However, most existing methods commit to a single such representation, or require complex full-scale decompositions, which makes them less practical for real-time use and limits their ability to generalize.

To address this, we introduce a \emph{lightweight complementary-cue fusion} framework. Our method extracts a compact Wavelet-Denoised Feature (WDF) channel that captures low-frequency structural information, and pairs it with a single complementary handcrafted cue that targets artifacts WDF cannot see: either local micro-texture (LBP) or up-sampling artifacts in the phase spectrum (SPSL). We propose two lightweight fusion variants:
\begin{itemize}
  \item \textbf{LFWL}: \emph{L}ightweight \emph{F}usion of \emph{W}DF with \emph{L}BP (low-frequency structure + local micro-texture)
  \item \textbf{LFWS}: \emph{L}ightweight \emph{F}usion of \emph{W}DF with \emph{S}PSL (low-frequency structure + phase)
\end{itemize}
In both variants, the two handcrafted maps are merged into a single additional channel before being concatenated with RGB, so the backbone always sees a four-channel input regardless of which complementary cue is used.

We show consistent improvements over single-cue baselines and earlier fusion methods on eight public benchmarks, including FaceForensics++, Celeb-DF v1/v2, and DFDC. Our approach achieves competitive cross-domain robustness with fewer parameters than comparable frequency-based detectors.
\section{Related Work}

\subsection{CNN-Based Detectors}
Early deepfake detectors used standard Convolutional Neural Networks, such as MesoNet \citep{afchar2018mesonet}, ResNet \citep{he2016deep}, EfficientNet-B4 \citep{tan2019efficientnet}, and Xception \citep{chollet2017xception}, which were trained on face-swap datasets. While these models performed well on their original data, they often struggled to adapt to new manipulation styles and datasets.

\subsection{Frequency-Aware Approaches}
Recent studies have moved away from using only raw RGB pixel values and now use specially designed frequency-based features to detect deepfakes.
F3Net fuses multiscale Fourier bands to expose synthesis errors \citep{qian2020thinking};  
SRM adopts steganalysis filters to highlight residual noise \citep{luo2021generalizing};  
SPSL keeps only the phase spectrum to expose up-sampling artifacts left by face-forgery pipelines \citep{liu2021spsl}.  
All three are benchmarked in DeepfakeBench \citep{yan2023deepfakebenchcomprehensivebenchmarkdeepfake} for reproducibility.

\subsection{Gap and Motivation}
A common pattern in the above detectors is that each one commits to a \emph{single} handcrafted representation: local micro-texture (LBP), spectral magnitude (F3Net, SRM), phase (SPSL), or low-frequency structure (wavelet methods).
Because deepfake artifacts can appear in any of these depending on the synthesis pipeline, single-cue detectors tend to overfit to dataset-specific artifacts and generalize poorly across manipulation styles.
A natural remedy is to combine cues, but full-scale multi-stream architectures inflate parameter count and complicate transfer to other backbones.
This motivates the central question of our work: can a \emph{minimal} learnable mixer of one low-frequency structural cue and one complementary cue (phase or local texture) recover most of the cross-domain robustness benefit at negligible cost?
We test this by combining a Wavelet-Denoised low-frequency map (WDF) \citep{mallat1989theory} with one complementary handcrafted cue, either LBP \citep{ojala2002multiresolution} (local micro-texture) or an SPSL-derived phase channel \citep{liu2021spsl}, via a single $1{\times}1$ convolution that adds only 292 parameters to the backbone.

\begin{figure*}[t]
  \centering

  \begin{minipage}[b]{0.45\textwidth}
    \centering
    \includegraphics[width=\linewidth]{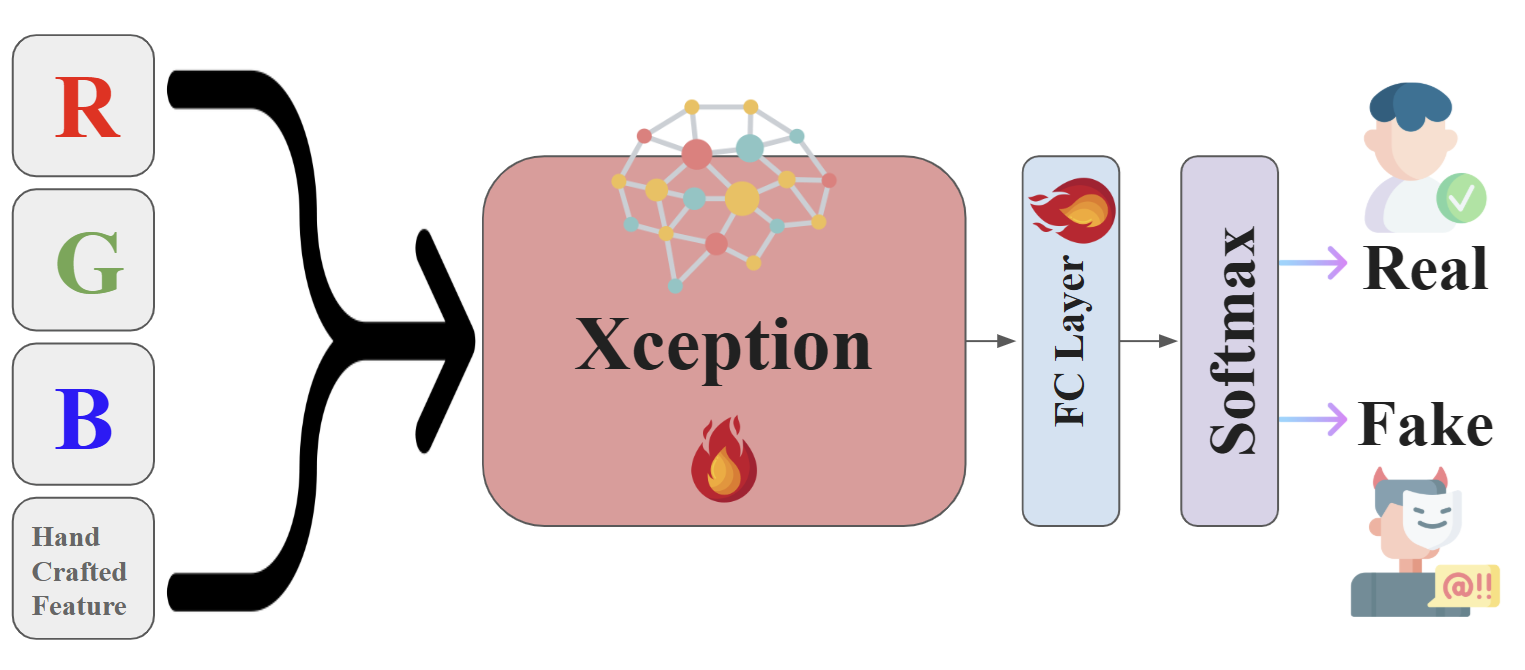}
    \vspace{0.5em}
    \textbf{Method 1: Concatenation Approach} \\
    One channel handcrafted feature (e.g., phase channel, LBP, WDF) is directly concatenated with the RGB image. The combined 4-channel input is passed to the backbone.
  \end{minipage}
  \hfill
  \begin{minipage}[b]{0.45\textwidth}
    \centering
    \includegraphics[width=\linewidth]{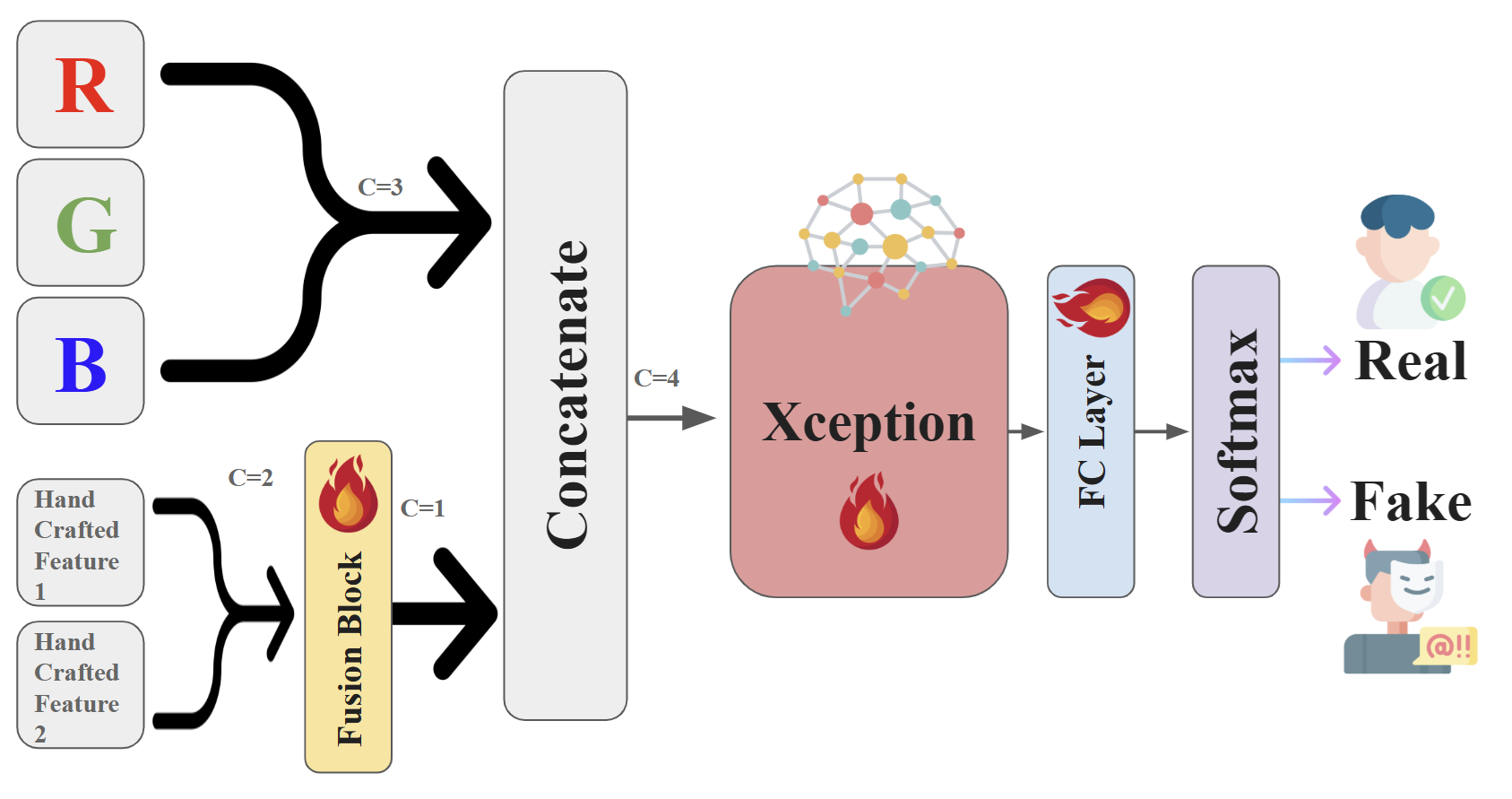}
    \vspace{0.5em}
    \textbf{Method 2: Learnable Fusion Block}\\ 
    A lightweight, trainable module that fuses multiple handcrafted feature streams (e.g., phase channel, WDF, LBP) into a single-channel representation, which is then concatenated with the RGB image.

  \end{minipage}

  \caption{Comparison of the two feature integration strategies. Method 1 naively concatenates a handcrafted feature map with RGB inputs, while Method 2 fuses selected feature maps into a compact representation, improving both accuracy and efficiency.}
  \label{fig:overview}
\end{figure*}

\section{Proposed Method}
\label{sec:Proposed Method}
\subsection{Overview of the Architecture}
Our proposed framework is built upon the Xception backbone to incorporate multiple handcrafted feature streams. As shown in Figure~\ref{fig:overview}, the model takes as input:
\begin{itemize}
    \item \textbf{RGB image}: normalized to the range \([-1, 1]\).
    \item \textbf{Handcrafted features}: extracted via various image-processing techniques, each also normalized to \([-1, 1]\).
\end{itemize}
We investigate two main strategies for combining these features with the RGB channels:
\begin{enumerate}
    \item \textbf{Method 1 (Concatenation):} Direct concatenation of the handcrafted feature maps and RGB image, fed into the Xception backbone.
    \item \textbf{Method 2 (Learnable Fusion Block):} A lightweight, trainable fusion block that fuses two feature streams into a single representation map before concatenation with the RGB image.
\end{enumerate}

All experiments are implemented on top of the DeepfakeBench framework \citep{yan2023deepfakebenchcomprehensivebenchmarkdeepfake}, which we extend to support both Method 1 (handcrafted features: LBP, WDF) and Method 2 (learnable fusion features: LFWS, LFWL). Our implementation enables consistent training and evaluation of both baseline and proposed models across a unified interface.

After training, the fusion block can be frozen, allowing the fused features alone to be transferred to alternative backbones (e.g., MesoNet, EfficientNet), thereby enabling flexibility and scalability.

\subsection{Handcrafted Feature maps and Normalization}
\label{sec:features}

We append three artefact-oriented channels to the RGB tensor:

\begin{itemize}
  \item \textbf{Wavelet-Denoised Feature (WDF).}  
        The grayscale frame is decomposed by a 3-level 2-D \texttt{db1} wavelet; all detail coefficients are zeroed and the low-frequency approximation is inverse–transformed, yielding a single coarse map that is re-scaled to $[-1,1]$.
    
    \item \textbf{LBP.}
    We compute \emph{uniform} Local Binary Patterns (LBP) with radius $1$ and $P{=}8$ neighbors using \texttt{local\_binary\_pattern} from \textsc{skimage}, resulting in codes ranging from $0$ to $P{+}1 = 9$. 
    To ensure stable, zero-centered inputs for CNN training, we linearly normalize the LBP codes to the range $[-1, 1]$. 
    The normalization is defined as
    \begin{equation}
        x_{\text{norm}} = \frac{x}{P+2} \times 2 - 1,
    \end{equation}
    where the denominator $P{+}2$ slightly compresses the upper bound, helping center the distribution closer to zero and improving training stability.

  \item \textbf{SPSL (phase channel).}
        Strictly speaking, SPSL~\citep{liu2021spsl} refers to a full detector that combines a spatial (RGB) stream with a phase-spectrum stream. In this work we reuse only the phase-spectrum input branch of that detector as a single handcrafted channel, which we refer to as the ``SPSL channel'' throughout for brevity.
        Following \citet{liu2021spsl}, we take the grayscale image, apply a 2-D FFT, keep only the phase spectrum ($\angle X$), set the magnitude to 1, and perform an inverse FFT. The resulting real part is a \emph{phase-only} image already bounded in $[-1,1]$.
\end{itemize}

Consequently, the WDF, LBP, and RGB channels are \emph{explicitly} normalised to
\([-1,1]\); the phase channel already lies in that range, so the
tensors\((R,G,B,\text{WDF},\text{LBP},\text{SPSL})\) feed the models
with balanced magnitudes.

\begin{figure}[t]
  \centering
  \includegraphics[width=0.5\linewidth]{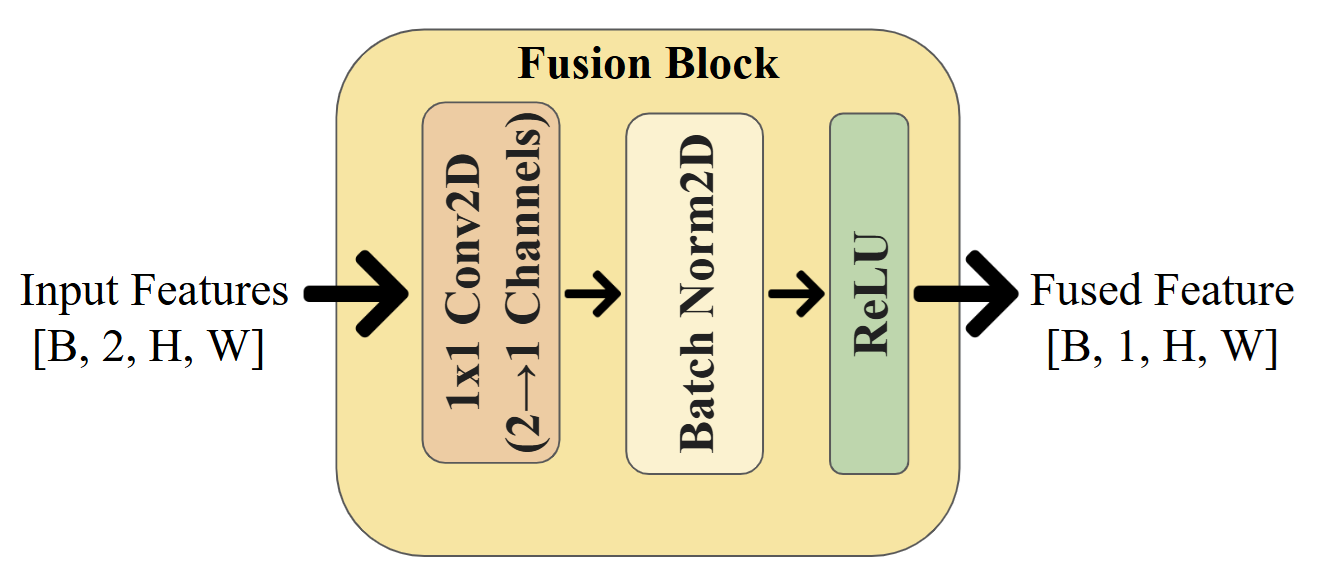}
  \vspace{0.5em}
  \caption{Architecture of the lightweight trainable fusion block. Two handcrafted channels (e.g., the phase channel and WDF) are concatenated along the channel dimension and passed through a $1\times1$ convolution, followed by batch normalization and ReLU. The output is a single fused feature map, which is then concatenated with the RGB image.}
  \label{fig:fusion_block}
\end{figure}
\subsection{Lightweight Fusion Block (Method 2)}
Figure~\ref{fig:fusion_block} illustrates the proposed module, which compresses two handcrafted maps into a single learned channel and concatenates it with RGB to form a 4-channel input for Xception.

\textbf{Design.} A $1{\times}1$ convolution mixes either WDF + phase channel (\textbf{LFWS}) or WDF + LBP (\textbf{LFWL}), followed by Batch Normalization and ReLU. This lightweight block introduces only two additional parameters per spatial location, stabilizes gradients, and couples the WDF low-frequency cue with a complementary cue (phase for LFWS, local micro-texture for LFWL).  

\textbf{Deployment.} After training, mixer weights and BN statistics are frozen, so the block can be reused with other backbones at negligible cost.  

\textbf{Advantages.} Reducing channels with Method 2 makes computation more efficient and helps the network learn a better representation instead of just combining features. Adding 292 extra weights, which is very small compared to the 21.9 million parameters in Xception (less than 0.0014 percent), leads to consistent accuracy improvements. This suggests that learnable pairwise fusion can detect important manipulation artifacts and reduce redundancy, which aligns with previous findings on channel efficiency.~\citep{hu2018senet,liang2020channel}.

\subsection{Implementation Details}

Below is a summary of the main training configuration parameters.

\begin{itemize}
    \item \textbf{Pretrained Backbone:} Xception initialized with the publicly available PyTorch weights used in DeepfakeBench~\citep{yan2023deepfakebenchcomprehensivebenchmarkdeepfake}.

    \item \textbf{Data Augmentation:} 
    Each transformation was applied independently with a probability of 0.5. These included horizontal flipping, random rotation (from -10° to +10°), Gaussian blur (kernel size 3 to 7), brightness and contrast adjustments (±10 \%), and JPEG compression (quality 40 to 100).

    \item \textbf{Normalization:} All channels (RGB and handcrafted feature maps) scaled to \([-1,1]\) using \(\mu=0.5\) and \(\sigma=0.5\).
    
    \item \textbf{Optimizer:} 
    We use the Adam optimizer with a learning rate of \(2\times10^{-4}\), 
    momentum parameters \(\beta_1=0.9\) and \(\beta_2=0.999\), 
    an epsilon of \(1\times10^{-8}\), 
    AMSGrad set to \texttt{false}, 
    and a weight decay of \(5\times10^{-4}\).

    \item \textbf{Training Schedule:} 
    We trained the model for 10 epochs using a batch size of 32, saving checkpoints after each epoch.

    \item \textbf{Loss Function:} 
    We use binary cross-entropy loss for all models except SRM, where Additive Margin Softmax (AM-Softmax) is applied to align with the original implementation.

    \item \textbf{Hardware:} 
    Training was performed on 2$\times$ NVIDIA V100 (32\,GB) GPUs.

    \item \textbf{Manual Seed:}
    All experiments were conducted with the random seed fixed to 1024.
    
\end{itemize}

\section{Experiments and Results}
\label{sec:experiments}

This section presents the experimental setup, datasets, and evaluation protocols, followed by qualitative and quantitative analyses of the proposed method. Unless specified otherwise, all implementation details, including optimizer, data augmentation, and normalization, are consistent with those described in Section~\ref{sec:Proposed Method}.

\subsection{Scope of Comparison}
Our study deliberately focuses on frequency-aware detectors and standard CNN backbones available in DeepfakeBench (Xception, EfficientNet-B4, ResNet-34, MesoNet, F3Net, SRM, SPSL). The central question we investigate is how complementary handcrafted cues can be combined within a classical CNN framework, so transformer-based, self-supervised, and disentanglement-based detectors are intentionally outside the scope of this work.

\subsection{Datasets \& Setup}

\textbf{Training.}
We train two identical networks under separate protocols to isolate the effect of training-data diversity on cross-domain robustness:
(i) \textbf{FaceForensics++ (c23)}~\citep{rossler2019faceforensics++}, the medium-compression version commonly used in benchmarks, which contains four classic manipulation styles (DF, F2F, FS, NT); and
(ii) \textbf{DFDC Preview (DFDCP)}~\citep{dolhansky2019dfdcpreview}, a $\sim$5\,k-video subset selected for its broader demographic and scene variation, testing generalization beyond familiar manipulations.
We do not train on the full Deepfake Detection Challenge (DFDC) dataset~\citep{dolhansky2020deepfake} ($>$100k videos) due to the compute cost of repeated experiments.

\textbf{Testing.}  
Both models are evaluated on eight public benchmarks: 
FaceForensics++ (c23)~\citep{rossler2019faceforensics++}, 
FaceShifter~\citep{li2020faceshifter}, 
Google DFD (DeepfakeDetection)~\citep{google2019dfd}, 
Celeb-DF v1 and v2~\citep{li2020celeb}, 
DFDC~\citep{dolhansky2020deepfake}, 
DFDCP~\citep{dolhansky2019dfdcpreview}, 
and UADFV~\citep{li2018ictuadfv}. 
 The average area under the curve (AUC) is reported across all benchmarks, without separating within-domain and cross-domain results. This unified metric incorporates both straightforward and complex cases, accounts for uncertainty in data provenance, and facilitates fair comparison by avoiding double-counting of datasets. 

All datasets are used with their official splits, ensuring a strict separation between training and testing; for instance, FF++ and DFDCP in the evaluation are drawn exclusively from their designated test sets. Each video is sampled into 32 frames, and train/test splits are made at the video level to prevent leakage, with models trained and evaluated on the extracted frames.

\begin{figure}[t]
  \centering
  \includegraphics[width=0.5\textwidth]{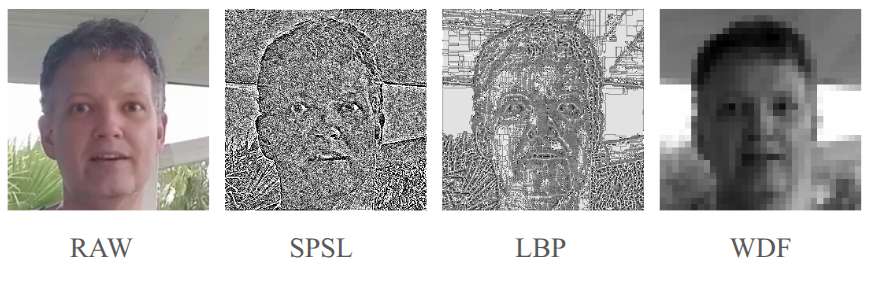}
  \caption{Visual examples of the handcrafted channels extracted from a single image. From left to right: original RGB, the phase-spectrum channel derived from SPSL, Local Binary Patterns (LBP), and Wavelet-Denoised Features (WDF). Each channel highlights different manipulation artifacts across spatial and frequency domains.}
  \label{fig:feature_visuals}
\end{figure}

\subsection{Baseline Analysis: Single-Channel Inputs}
We evaluate the discriminative power of each handcrafted channel by feeding individual channels, Wavelet-Denoised Features (WDF), the SPSL-derived phase channel, and Local Binary Patterns (LBP), into an Xception backbone without concatenating raw RGB images.

\begin{itemize}
\item \textbf{WDF:} AUC = 0.9519
\item \textbf{Phase channel (SPSL):} AUC = 0.9477
\item \textbf{LBP:} AUC = 0.9388
\end{itemize}
WDF achieves the highest AUC, showing that resampling and compression artifacts provide informative signals in low-frequency regions. The phase channel and LBP also yield strong performance, supporting the assertion that phase and local-texture cues encode relevant information for forgery detection. These findings provide a rationale for fusing complementary representations, such as WDF with the phase channel or LBP, to capture a broader spectrum of artifacts. These baseline models were trained and evaluated on the FaceForensics++ dataset without data augmentation.

\subsection{Fusion Approaches and Final Detection Performance}

There are two fusion strategies, as detailed in Section~\ref{sec:Proposed Method} and illustrated in Figure~\ref{fig:overview}. Method 1 utilizes direct concatenation, while Method 2 employs a lightweight fusion block. For Method 2, two fused models are evaluated: LFWS, which combines WDF and the phase channel, and LFWL, which combines WDF and LBP.

The baseline models consist of standard convolutional neural networks (CNNs), including Xception, Meso-4 (MesoNet), ResNet-34, and EfficientNet-B4, as well as frequency-based models such as F3Net, SRM, and SPSL. Although SPSL utilizes feature concatenation as in Method~1, it is categorized with frequency-based models to maintain consistency with the original implementation and ensure fair comparison.

All handcrafted feature maps are normalized or rescaled to the range [-1,1] prior to fusion or concatenation. This preprocessing step ensures balanced training across feature representations.

\subsubsection{Trained on FaceForensics++ (c23)}

\paragraph{Overall Results.}
Table~\ref{tab:method_ffpp} reports AUC scores across eight benchmarks for models trained solely on FaceForensics++ (c23). The results include baseline CNNs, handcrafted feature-based variants, frequency-based models (F3Net, SRM, SPSL), and our proposed approaches (Method 1: concatenation; Method 2: lightweight fusion).

\noindent
\textbf{Key Observations (FF++ Training).}
\begin{itemize}
    \item \textbf{Fusion methods generalize best.}  
    Among all models, Method 2 fusions demonstrate the strongest generalization. LFWS (fusing WDF + phase channel) achieves the highest average AUC of \textbf{\textcolor{blue}{0.7858}}, outperforming every baseline and frequency-based method. LFWL (WDF + LBP) also ranks second with \textbf{\textcolor{blue}{0.7786}}, followed closely by WDF (\textbf{\textcolor{blue}{0.7708}}). These results show the complementary nature of handcrafted features when fused with a learnable module.
    
    \item \textbf{Notable improvements on challenging datasets.}  
    On cross-domain benchmarks, including Celeb-DF-v1 and Celeb-DF-v2, LFWS achieves AUC scores of 0.7875 and 0.7548, respectively. These values are significantly higher than those of all baseline models, including SRM and F3Net. The results indicate that the fused representations provide increased robustness to dataset shifts and previously unseen manipulations.
    
    \item \textbf{Method 2 outperforms Method 1.}  
    Direct concatenation in Method 1 yields only modest improvements over the baselines and does not achieve the performance of Method 2. This performance gap demonstrates the benefit of a lightweight, trainable fusion block that compresses and non-linearly refines feature interactions, rather than simply stacking them.

    \item \textbf{Baseline CNNs and frequency-based methods plateau.}  
    While SRM (\(0.7672\)) and F3Net (\(0.7689\)) perform competitively, they do not surpass the handcrafted + fusion combinations. Among CNN baselines, EfficientNet-B4 (\(0.7621\)) is the best, though it still falls short of the fusion-based approaches.
\end{itemize}

\begin{table*}[t]
\centering
\footnotesize
\setlength{\tabcolsep}{3.5pt}
\renewcommand{\arraystretch}{1.15}
\caption{\textbf{Forgery detection AUC trained on FF++ (c23).}
Top-3 average AUC values are highlighted in \textbf{\textcolor{topAUC}{deep blue}}.
The bottom row reports the relative change (\%) in average AUC over the Xception baseline,
illustrating the benefit of integrating handcrafted and fused features for cross-domain robustness.}
\label{tab:method_ffpp}
\begin{tabular}{l YYYY YYY YY YY}
\toprule
& \multicolumn{4}{c}{\cellcolor{grpA}\textbf{Baseline CNNs}}
& \multicolumn{3}{c}{\cellcolor{grpB}\textbf{Frequency-based}}
& \multicolumn{2}{c}{\cellcolor{grpC}\textbf{Method 1}}
& \multicolumn{2}{c}{\cellcolor{grpD}\textbf{Method 2: Fusion}} \\
\cmidrule(lr){2-5}\cmidrule(lr){6-8}\cmidrule(lr){9-10}\cmidrule(lr){11-12}
\textbf{Dataset}
& {\textbf{Xcept}} & {\textbf{Meso4}} & {\textbf{RNet34}} & {\textbf{EffB4}}
& {\textbf{F3Net}} & {\textbf{SRM}}  & {\textbf{SPSL}}
& {\textbf{LBP}}   & {\textbf{WDF}}
& {\textbf{LFWS}}  & {\textbf{LFWL}} \\
\midrule
FaceForensics++ & 0.9805 & 0.7135 & 0.9665 & 0.9765 & 0.9791 & 0.9770 & 0.9789 & 0.9662 & 0.9778 & 0.9761 & 0.9741 \\
\midrule
FaceShifter     & 0.6248 & 0.6275 & 0.5506 & 0.6062 & 0.6062 & 0.5663 & 0.6629 & 0.6434 & 0.6581 & 0.6163 & 0.6150 \\
DFD             & 0.8042 & 0.5524 & 0.7691 & 0.8535 & 0.8281 & 0.8111 & 0.8141 & 0.7983 & 0.8088 & 0.8241 & 0.8146 \\
Celeb-DF-v1     & 0.6238 & 0.6022 & 0.6607 & 0.7363 & 0.7052 & 0.7466 & 0.6734 & 0.7106 & 0.6860 & 0.7875 & 0.7188 \\
Celeb-DF-v2     & 0.6875 & 0.6080 & 0.6712 & 0.6684 & 0.7083 & 0.7345 & 0.6999 & 0.7216 & 0.7108 & 0.7548 & 0.7324 \\
DFDCP           & 0.6764 & 0.5734 & 0.6355 & 0.6737 & 0.7259 & 0.7019 & 0.7249 & 0.7149 & 0.7244 & 0.6991 & 0.7345 \\
DFDC            & 0.6862 & 0.5720 & 0.6932 & 0.6747 & 0.6938 & 0.6838 & 0.7024 & 0.6940 & 0.6838 & 0.7172 & 0.7072 \\
UADFV           & 0.8966 & 0.8714 & 0.9099 & 0.9072 & 0.9043 & 0.9163 & 0.8805 & 0.8975 & 0.9167 & 0.9110 & 0.9325 \\
\midrule
\rowcolor{rowAvg}
\textbf{Avg AUC}
& \bfseries 0.7475 & \bfseries 0.6401 & \bfseries 0.7321 & \bfseries 0.7621
& \bfseries 0.7689 & \bfseries 0.7672 & \bfseries 0.7671
& \bfseries 0.7683 & {\bfseries\color{topAUC} 0.7708}
& {\bfseries\color{topAUC} 0.7858} & {\bfseries\color{topAUC} 0.7786} \\
\rowcolor{rowImp}
\textbf{$\Delta$ vs.\ Xcept (\%)}
& \multicolumn{1}{c}{\bfseries --}
& \multicolumn{1}{c}{\bfseries $-$10.74}
& \multicolumn{1}{c}{\bfseries $-$1.54}
& \multicolumn{1}{c}{\bfseries $+$1.46}
& \multicolumn{1}{c}{\bfseries $+$2.14}
& \multicolumn{1}{c}{\bfseries $+$1.97}
& \multicolumn{1}{c}{\bfseries $+$1.96}
& \multicolumn{1}{c}{\bfseries $+$2.08}
& \multicolumn{1}{c}{\bfseries $+$2.33}
& \multicolumn{1}{c}{\bfseries $+$3.83}
& \multicolumn{1}{c}{\bfseries $+$3.11} \\
\bottomrule
\end{tabular}
\end{table*}

\subsubsection{Trained on DFDCP}

\paragraph{Overall Results.}  
Table \ref{tab:method_dfdcp} lists AUC scores for models trained on DFDCP (same number of videos as FF++ but broader real-world variability) and evaluated on eight benchmarks.

\noindent
\textbf{Analysis (DFDCP Training).}
\begin{itemize}
    \item \textbf{Fusion methods generalize best.}  
    Method 2 again produces the two strongest detectors: LFWS (\textcolor{blue}{0.7495}), LFWL (\textcolor{blue}{0.7470}).

    \item \textbf{Single-cue gains shrink, fusion gains grow.}
    With changed dataset from FF++ to DFDCP, the relative improvements of F3Net/SRM change to {\small–1.42 \%}/{\small+1.59 \%}, whereas LFWS/LFWL rise to {\small+4.44 \%}/{\small+4.19 \%}.  This indicates that single-cue detectors overfit to the specific artifacts present in the training set, whereas the \emph{learnable fusion block} still discovers complementary structural and phase/texture evidence that transfers.

    \item \textbf{Superior cross-domain robustness.}
    Although individual frequency-based baselines outperform the fusion models on certain datasets, the fusion block delivers \emph{the highest \textbf{average} AUC across all tests}. This overall lead suggests that our method captures manipulation-invariant features that generalize across demographic variations and previously unseen forgery styles, rather than overfitting to dataset-specific artifacts.
\end{itemize}

\noindent
When training datasets are matched in size but differ in style, the LFWS and LFWL models maintain or increase their performance advantage. This result demonstrates robustness that does not depend on the statistics of any single dataset.

\begin{table*}[t]
\centering
\footnotesize
\setlength{\tabcolsep}{3.5pt}
\renewcommand{\arraystretch}{1.15}
\caption{\textbf{Forgery detection AUC trained on DFDCP.}
Top-3 average AUC values are highlighted in \textbf{\textcolor{topAUC}{deep blue}}.
The bottom row reports the relative change (\%) in average AUC over the Xception baseline,
illustrating the benefit of integrating handcrafted and fused features for cross-domain robustness.}
\label{tab:method_dfdcp}
\begin{tabular}{l YYYY YYY YY YY}
\toprule
& \multicolumn{4}{c}{\cellcolor{grpA}\textbf{Baseline CNNs}}
& \multicolumn{3}{c}{\cellcolor{grpB}\textbf{Frequency-based}}
& \multicolumn{2}{c}{\cellcolor{grpC}\textbf{Method 1}}
& \multicolumn{2}{c}{\cellcolor{grpD}\textbf{Method 2: Fusion}} \\
\cmidrule(lr){2-5}\cmidrule(lr){6-8}\cmidrule(lr){9-10}\cmidrule(lr){11-12}
\textbf{Dataset}
& {\textbf{Xcept}} & {\textbf{Meso4}} & {\textbf{RNet34}} & {\textbf{EffB4}}
& {\textbf{F3Net}} & {\textbf{SRM}}  & {\textbf{SPSL}}
& {\textbf{LBP}}   & {\textbf{WDF}}
& {\textbf{LFWS}}  & {\textbf{LFWL}} \\
\midrule
DFDCP            & 0.9408 & 0.7937 & 0.9411 & 0.9313 & 0.9245 & 0.9380 & 0.9209 & 0.9213 & 0.9168 & 0.9262 & 0.9224 \\
\midrule
DFDC             & 0.6744 & 0.5729 & 0.6334 & 0.6890 & 0.6612 & 0.6856 & 0.6809 & 0.6713 & 0.6882 & 0.7061 & 0.6716 \\
FaceForensics++  & 0.6295 & 0.6132 & 0.6074 & 0.6462 & 0.6264 & 0.6404 & 0.6499 & 0.6274 & 0.6556 & 0.6464 & 0.6457 \\
FaceShifter      & 0.4932 & 0.6246 & 0.4990 & 0.5419 & 0.5108 & 0.5237 & 0.5211 & 0.5519 & 0.5305 & 0.5561 & 0.5761 \\
DFD              & 0.7401 & 0.6453 & 0.7004 & 0.7610 & 0.7062 & 0.7812 & 0.7515 & 0.7065 & 0.7294 & 0.7688 & 0.7385 \\
Celeb-DF-v1      & 0.6996 & 0.6504 & 0.6066 & 0.6583 & 0.6497 & 0.6486 & 0.6565 & 0.7134 & 0.7181 & 0.7488 & 0.7659 \\
Celeb-DF-v2      & 0.6873 & 0.6737 & 0.6928 & 0.6956 & 0.6812 & 0.7188 & 0.6797 & 0.7180 & 0.7297 & 0.7539 & 0.7651 \\
UADFV            & 0.7757 & 0.8451 & 0.7710 & 0.8000 & 0.7669 & 0.8317 & 0.7937 & 0.8676 & 0.8749 & 0.8896 & 0.8903 \\
\midrule
\rowcolor{rowAvg}
\textbf{Avg AUC}
& \bfseries 0.7051 & \bfseries 0.6774 & \bfseries 0.6815 & \bfseries 0.7154
& \bfseries 0.6909 & \bfseries 0.7210 & \bfseries 0.7068
& \bfseries 0.7222 & {\bfseries\color{topAUC} 0.7304}
& {\bfseries\color{topAUC} 0.7495} & {\bfseries\color{topAUC} 0.7470} \\
\rowcolor{rowImp}
\textbf{$\Delta$ vs.\ Xcept (\%)}
& \multicolumn{1}{c}{\bfseries --}
& \multicolumn{1}{c}{\bfseries $-$2.77}
& \multicolumn{1}{c}{\bfseries $-$2.36}
& \multicolumn{1}{c}{\bfseries $+$1.03}
& \multicolumn{1}{c}{\bfseries $-$1.42}
& \multicolumn{1}{c}{\bfseries $+$1.59}
& \multicolumn{1}{c}{\bfseries $+$0.17}
& \multicolumn{1}{c}{\bfseries $+$1.71}
& \multicolumn{1}{c}{\bfseries $+$2.53}
& \multicolumn{1}{c}{\bfseries $+$4.44}
& \multicolumn{1}{c}{\bfseries $+$4.19} \\
\bottomrule
\end{tabular}
\end{table*}

\subsection{Ablation Study: Frozen Fusion Block Transferability}
\label{subsec:frozen_block}

\begin{figure}[tb]
  \centering
  \includegraphics[width=0.5\linewidth]{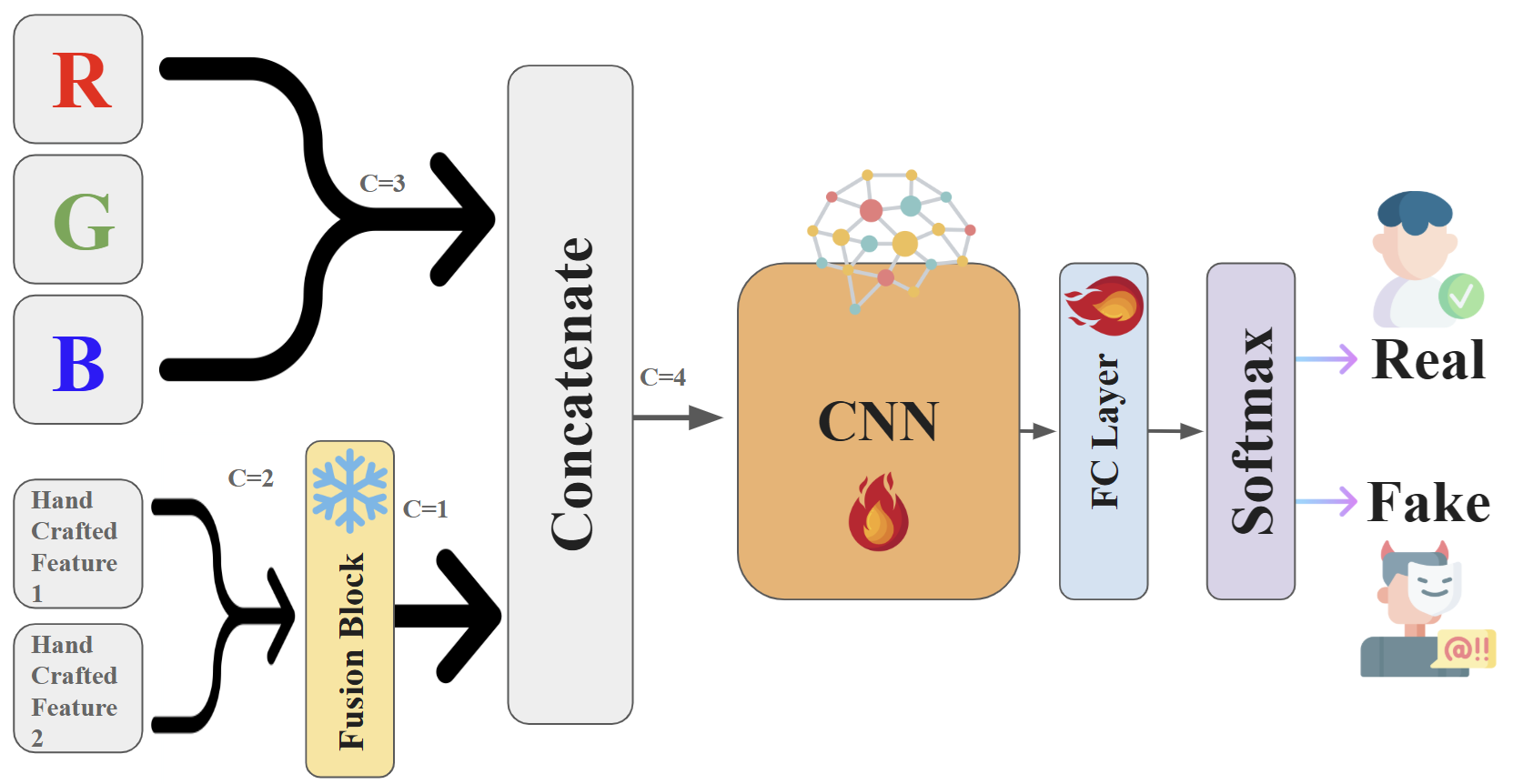}
  \caption{Frozen Fusion Block Inference Setup. The fusion block (trained with Xception on FF++) is frozen and reused as a preprocessing step for other backbones.}
  \label{fig:frozen_fusion_block}
\end{figure}

We train LFWS and LFWL with Xception on FaceForensics++ (c23), freeze the fusion block, and attach it, without any retraining, to Meso-4 (MesoNet), ResNet-34, and EfficientNet-B4 (Tab.~\ref{tab:auc_comparison_all}; Fig.~\ref{fig:frozen_fusion_block}).

\paragraph{Backbone-wise analysis.}
The frozen fusion block trained on Xception is reused directly, isolating its cross-backbone generalization. While both LFWS and LFWL consistently improve performance across backbones (see Tab.~\ref{tab:auc_comparison_all}), we focus on the LFWS results for clarity. Xception achieves the highest improvement, with an increase of 3.8 in area under the curve (AUC). The frozen block processes feature statistics identical to those present during training.
ResNet-34 demonstrates a 2.7 increase in AUC. The initial 3×3 convolutional layers and residual connections maintain the integrated feature representation, facilitating effective block transfer.
EfficientNet-B4 shows a 0.9 increase in AUC. Because squeeze-and-excitation (SE) layers already perform channel re-weighting~\citep{hu2018senet}, further gains are small and only matter when accuracy changes of less than 1\% are significant.
Meso-4 (MesoNet) achieves a 1.0 increase in AUC. While aggressive down-sampling reduces the effect of the extra channel, adding 292 parameters still helps edge models with fewer than five million parameters.

The frozen fusion block is transferable across architectures, with the largest relative gains observed on mid-depth convolutional neural networks such as Xception and ResNet-34. These networks retain sufficient capacity to utilize an additional handcrafted channel without significantly increasing model size.

\begin{figure}[tb]
  \centering
  \includegraphics[width=0.5\linewidth]{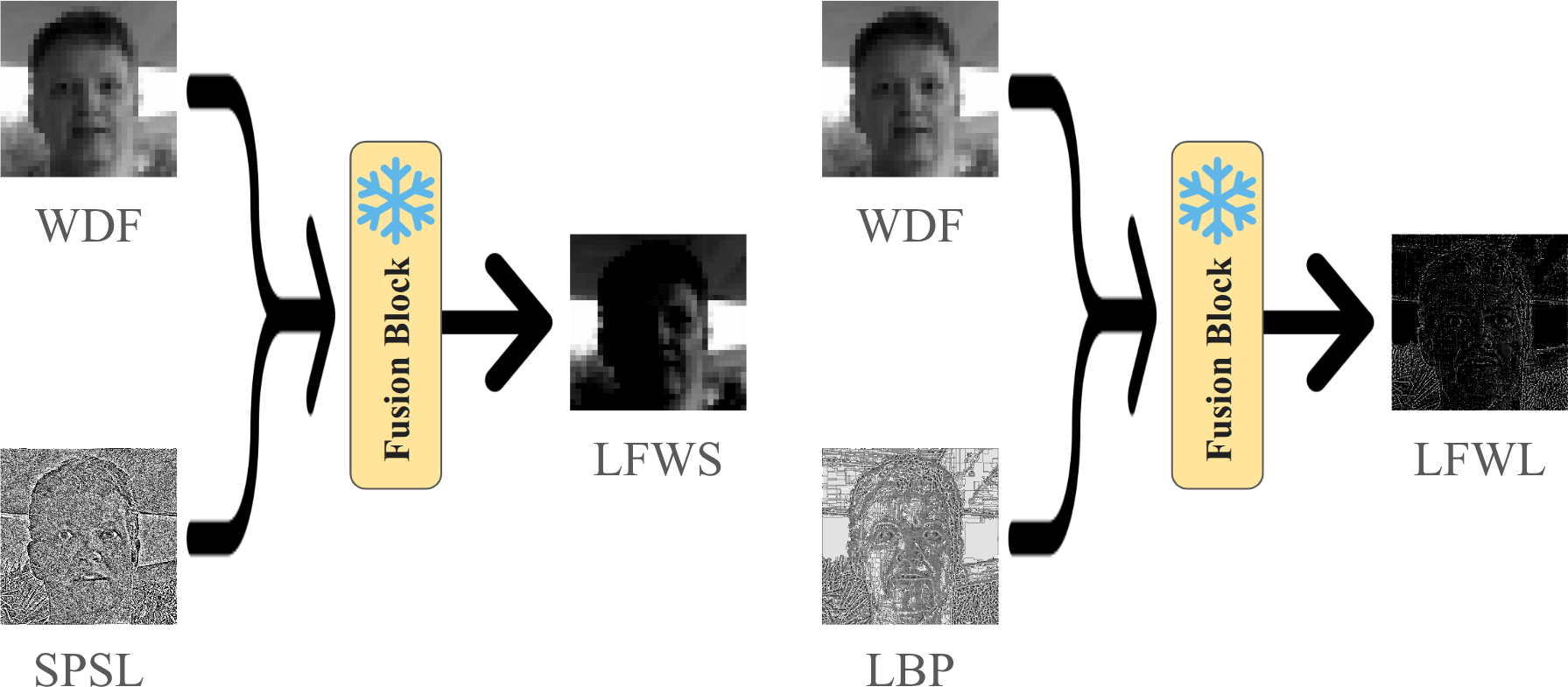}
  \caption{Learned mixing weights of the $1{\times}1$ convolution in our lightweight fusion block.
  Each variant fuses two handcrafted streams into a single channel using two scalar weights, one per input stream.
  LFWS assigns $+0.18$ to WDF and $-0.12$ to the phase channel; LFWL assigns $+0.08$ to WDF and $-0.32$ to LBP.
  Since none of the four learned weights are zero, the block keeps both streams in every variant rather than collapsing onto a single one.
  The opposite signs further indicate that the two streams are combined as complementary (additive vs.\ subtractive) cues, not as redundant copies of the same signal.}
  \label{fig:fused_features}
\end{figure}

\begin{table*}[t]
\centering
\scriptsize
\setlength{\tabcolsep}{3pt}
\renewcommand{\arraystretch}{1.05}
\caption{\textbf{AUC comparison across alternative backbones.}
Baseline backbones without additional features are included for reference.
The bottom two rows summarize the average AUC across all datasets and the relative change (\%)
over each backbone's baseline, demonstrating the added value of incorporating handcrafted features
through the (frozen) fusion block.}
\label{tab:auc_comparison_all}
\begin{tabular}{l YYYY YYYY YYYY}
\toprule
& \multicolumn{4}{c}{\cellcolor{grpA}\textbf{Baseline}}
& \multicolumn{4}{c}{\cellcolor{grpC}\textbf{LFWL (frozen)}}
& \multicolumn{4}{c}{\cellcolor{grpD}\textbf{LFWS (frozen)}} \\
\cmidrule(lr){2-5}\cmidrule(lr){6-9}\cmidrule(lr){10-13}
\textbf{Dataset}
& {\textbf{Xcept}} & {\textbf{EffB4}} & {\textbf{RNet34}} & {\textbf{Meso4}}
& {\textbf{Xcept}} & {\textbf{EffB4}} & {\textbf{RNet34}} & {\textbf{Meso4}}
& {\textbf{Xcept}} & {\textbf{EffB4}} & {\textbf{RNet34}} & {\textbf{Meso4}} \\
\midrule
FaceForensics++ & 0.9805 & 0.9765 & 0.9665 & 0.7135 & 0.9741 & 0.9787 & 0.9605 & 0.6922 & 0.9761 & 0.9801 & 0.9616 & 0.6653 \\
\midrule
FaceShifter     & 0.6248 & 0.6062 & 0.5506 & 0.6275 & 0.6150 & 0.6002 & 0.5986 & 0.5904 & 0.6163 & 0.6381 & 0.6006 & 0.6031 \\
DFD             & 0.8042 & 0.8535 & 0.7691 & 0.5524 & 0.8146 & 0.8387 & 0.7996 & 0.5671 & 0.8241 & 0.8385 & 0.7699 & 0.6235 \\
Celeb-DF-v1     & 0.6238 & 0.7363 & 0.6607 & 0.6022 & 0.7188 & 0.7262 & 0.7836 & 0.5383 & 0.7875 & 0.7238 & 0.7830 & 0.6310 \\
Celeb-DF-v2     & 0.6875 & 0.6684 & 0.6712 & 0.6080 & 0.7324 & 0.7430 & 0.7638 & 0.6135 & 0.7548 & 0.6958 & 0.7238 & 0.6175 \\
DFDCP           & 0.6764 & 0.6737 & 0.6355 & 0.5734 & 0.7345 & 0.7078 & 0.6752 & 0.6744 & 0.6991 & 0.6753 & 0.6639 & 0.6255 \\
DFDC            & 0.6862 & 0.6747 & 0.6932 & 0.5720 & 0.7072 & 0.6980 & 0.6610 & 0.5886 & 0.7172 & 0.6930 & 0.6527 & 0.5701 \\
UADFV           & 0.8966 & 0.9072 & 0.9099 & 0.8714 & 0.9325 & 0.9454 & 0.8983 & 0.8691 & 0.9110 & 0.9205 & 0.9149 & 0.8637 \\
\midrule
\rowcolor{rowAvg}
\textbf{Avg AUC}
& \bfseries 0.7475 & \bfseries 0.7621 & \bfseries 0.7321 & \bfseries 0.6401
& \bfseries 0.7786 & \bfseries 0.7798 & \bfseries 0.7676 & \bfseries 0.6417
& \bfseries 0.7858 & \bfseries 0.7706 & \bfseries 0.7588 & \bfseries 0.6500 \\
\rowcolor{rowImp}
\textbf{$\Delta$ vs.\ Baseline (\%)}
& \multicolumn{1}{c}{\bfseries --}
& \multicolumn{1}{c}{\bfseries --}
& \multicolumn{1}{c}{\bfseries --}
& \multicolumn{1}{c}{\bfseries --}
& \multicolumn{1}{c}{\bfseries $+$3.11}
& \multicolumn{1}{c}{\bfseries $+$1.77}
& \multicolumn{1}{c}{\bfseries $+$3.55}
& \multicolumn{1}{c}{\bfseries $+$0.16}
& \multicolumn{1}{c}{\bfseries $+$3.83}
& \multicolumn{1}{c}{\bfseries $+$0.85}
& \multicolumn{1}{c}{\bfseries $+$2.67}
& \multicolumn{1}{c}{\bfseries $+$0.99} \\
\bottomrule
\end{tabular}
\end{table*}

\section{Discussion}

\paragraph{Lightweight fusion vs.\ direct concatenation.}
Appending handcrafted feature maps directly to RGB (Method~1) already raises AUC over a vanilla CNN, but our lightweight block (Method~2) consistently does better while adding fewer parameters. The key difference is that Method~2 first \emph{compresses} two handcrafted maps into a single learned channel and then concatenates it with RGB, preserving the original 4-channel input shape. Across eight benchmarks, Method~2 delivers the highest \emph{average} AUC, indicating that learned mixing of complementary cues is superior to na\"ive stacking. Both fusion variants (LFWL: WDF + LBP; LFWS: WDF + phase channel) surpass their Method~1 counterparts, confirming that the gain stems from better feature synergy rather than model capacity.

\paragraph{Why does fusion generalize better than single-cue detectors?}
A useful lens is the change in the relative improvement of single-cue baselines vs.\ fusion when the training set shifts from FF++ to DFDCP. F3Net's relative gain drops from $+2.14\%$ to $-1.42\%$ and SRM's from $+1.97\%$ to $+1.59\%$, whereas LFWS/LFWL rise from $+3.83\%/+3.11\%$ to $+4.44\%/+4.19\%$. Single-cue detectors appear to latch onto artifacts specific to the training set, while the lightweight fusion block continues to discover complementary evidence (low-frequency structure combined with either phase or local micro-texture) that transfers across domains.

\begin{figure}[t]
  \centering
  \includegraphics[width=0.95\linewidth]{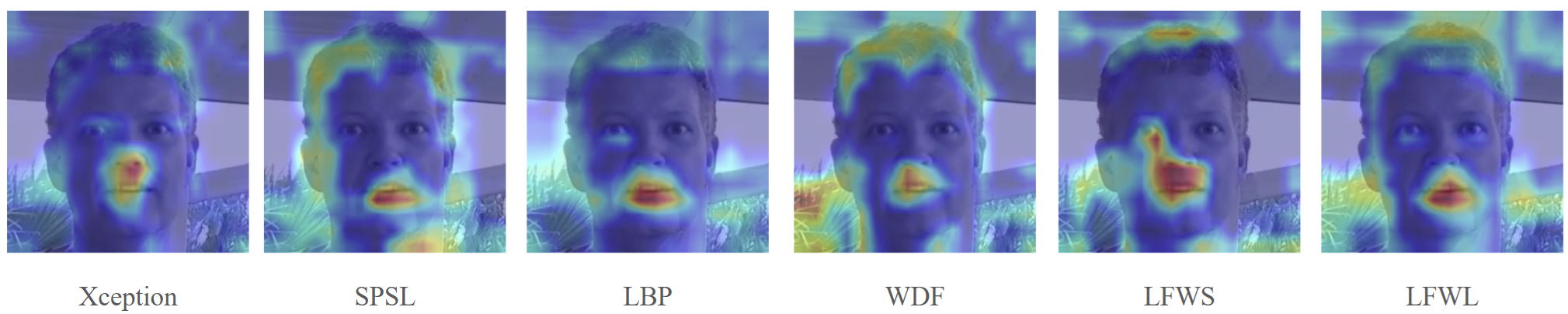}
  \caption{
    Grad-CAM heatmaps on a forged frame. Xception focuses broadly on central facial regions, while SPSL and LBP highlight phase and texture inconsistencies near the jaw and lips. WDF emphasizes boundary and hairline areas where low-frequency blending artifacts often occur. Fusion-based models (LFWS, LFWL) yield more comprehensive attention across discriminative regions.
  }
  \label{fig:gradcam}
\end{figure}

\paragraph{Where do the fused models look? (Grad-CAM)}
Figure~\ref{fig:gradcam} shows model attention on a forged frame. Xception attends broadly to the nose and mouth; SPSL focuses on phase artifacts near the jaw; LBP highlights fine-grained textures around the lips; and WDF concentrates on the face boundary and hairline, where deepfake blending often introduces lighting and color inconsistencies. The fusion variants (LFWS, LFWL) yield more balanced and focused attention across these discriminative regions, suggesting that the $1{\times}1$ mixer integrates rather than averages the underlying cues.

\paragraph{Transferability of the frozen block.}
The frozen-block experiment (Section~\ref{subsec:frozen_block}) further indicates that the mixer learns architecture-agnostic statistics: a block trained once with Xception on FF++ lifts AUC on Meso-4, ResNet-34, and EfficientNet-B4 without any fine-tuning. The largest relative gains appear in mid-depth networks (Xception, ResNet-34), which have enough capacity to exploit the extra channel without the diminishing returns observed on deeper EfficientNet-B4. A single, once-trained mixer therefore generalizes broadly across both datasets and backbones.

\section{Conclusion}
We presented a lightweight fusion block that combines two handcrafted streams, WDF together with either LBP (LFWL) or the SPSL-derived phase channel (LFWS), into a single learned channel that is concatenated with RGB before being passed to a standard CNN backbone. The block adds only 292 parameters to a 21.9M-parameter Xception ($<\!0.0014\%$ overhead), keeping the overall model smaller than F3Net (22.5M) and less than half the size of SRM (55.3M).

Despite this minimal cost, the two fused detectors achieve the highest average AUC among the eight detectors compared on eight public benchmarks, raising AUC from 74.8\% to 78.6\% when trained on FaceForensics++ and from 70.5\% to 74.9\% when trained on DFDC-Preview (gains of $+3.8\%$ and $+4.4\%$ over the Xception baseline). The block is also transferable: once trained with Xception, the frozen mixer improves Meso-4, ResNet-34, and EfficientNet-B4 with no retraining.

Taken together, these results suggest that carefully paired handcrafted features, combined through a minimal learnable mixer, can match or exceed the cross-domain robustness of much larger frequency-based detectors. We see this as evidence for reconsidering scale-driven design choices in face video forgery detection and as a starting point for richer but still lightweight fusion strategies, including more than two streams, attention-based mixing, and learned per-pixel weights.

\section{Limitations}
We note several limitations of this study.
\textbf{(i) Single-seed runs.} All results are reported from a single training run with a fixed random seed (1024); we did not measure run-to-run variance, so reported gains should be interpreted as point estimates rather than statistically validated effects.
\textbf{(ii) Restricted comparison set.} We deliberately focus on frequency-aware detectors and standard CNN backbones available in DeepfakeBench, since the central question is how complementary handcrafted cues can be combined within a classical CNN framework. We do not compare against transformer-based, self-supervised, or disentanglement-based methods (e.g., UCF); claims about competitiveness should be read relative to this comparison set.
\textbf{(iii) Training-set scale.} We train on FaceForensics++ (c23) and DFDC-Preview rather than the full DFDC dataset, due to compute constraints, which may limit the strength of generalization claims.
\textbf{(iv) Pairwise fusion only.} Our fusion is restricted to pairwise combinations of WDF (low-frequency structure) with one complementary cue (phase or local micro-texture); whether richer fusion strategies (more streams, attention-based mixing, learned per-pixel weights) would yield further gains is left to future work.

\newpage

\bibliography{references}

@inproceedings{rossler2019faceforensics++,
  title     = {FaceForensics++: Learning to Detect Manipulated Facial Images},
  author    = {R{\"o}ssler, Andreas and Cozzolino, Davide and Verdoliva, Luisa and Riess, Christian and Thies, Justus and Nie{\ss}ner, Matthias},
  booktitle = {Proceedings of the IEEE/CVF International Conference on Computer Vision (ICCV)},
  year      = {2019},
  pages     = {1--11}
}

@article{dolhansky2019dfdcpreview,
  title   = {The Deepfake Detection Challenge (DFDC) Preview Dataset},
  author  = {Dolhansky, Brian and Ferrer, Cristian and Bitton, Michael and Pflaum, Ben and Baram, Jikuo and Rohrbach, Marcus},
  journal = {arXiv preprint arXiv:1910.08854},
  year    = {2019},
  url     = {https://arxiv.org/abs/1910.08854}
}

@inproceedings{liu2021spsl,
  title     = {Spatial-Phase Shallow Learning: Rethinking Face Forgery Detection in Frequency Domain},
  author    = {Liu, Honggu and Li, Xiaodan and Zhou, Wenbo and Chen, Yuefeng and He, Yuan and Xue, Hui and Zhang, Weiming and Yu, Nenghai},
  booktitle = {Proceedings of the IEEE/CVF Conference on Computer Vision and Pattern Recognition (CVPR)},
  year      = {2021},
  pages     = {772--781}
}

@inproceedings{qian2020thinking,
  title     = {Thinking in Frequency: Face Forgery Detection by Mining Frequency-Aware Clues},
  author    = {Qian, Yuyang and Yin, Guojun and Sheng, Lu and Chen, Zixuan and Shao, Jing},
  booktitle = {European Conference on Computer Vision (ECCV)},
  pages     = {86--103},
  year      = {2020},
  publisher = {Springer}
}

@inproceedings{luo2021generalizing,
  title     = {Generalizing Face Forgery Detection with High-Frequency Features},
  author    = {Luo, Yuchen and Zhang, Yong and Yan, Junchi and Liu, Wei},
  booktitle = {Proceedings of the IEEE/CVF Conference on Computer Vision and Pattern Recognition (CVPR)},
  year      = {2021}
}

@misc{yan2023deepfakebenchcomprehensivebenchmarkdeepfake,
  title        = {DeepfakeBench: A Comprehensive Benchmark of Deepfake Detection}, 
  author       = {Zhiyuan Yan and Yong Zhang and Xinhang Yuan and Siwei Lyu and Baoyuan Wu},
  year         = {2023},
  eprint       = {2307.01426},
  archivePrefix= {arXiv},
  primaryClass = {cs.CV},
  url          = {https://arxiv.org/abs/2307.01426},
  note         = {\url{https://github.com/SCLBD/DeepfakeBench}}
}

@inproceedings{wolter2021waveletpackets,
  title     = {Wavelet-Packets for Deepfake Image Analysis and Detection},
  author    = {Wolter, Moritz and Blanke, Felix and Heese, Raoul and Garcke, Jochen},
  booktitle = {International Conference on Artificial Neural Networks (ICANN)},
  year      = {2021},
  note      = {arXiv:2106.09369}
}

@inproceedings{dzanic2020fourier,
  title     = {Fourier Spectrum Discrepancies in Deep Network Generated Images},
  author    = {Dzanic, Tarik and Shah, Karan and Witherden, Freddie},
  booktitle = {Advances in Neural Information Processing Systems (NeurIPS)},
  volume    = {33},
  pages     = {3022--3032},
  year      = {2020}
}

@inproceedings{schwarz2021frequency,
  title     = {On the Frequency Bias of Generative Models},
  author    = {Schwarz, Katja and Liao, Yiyi and Geiger, Andreas},
  booktitle = {Advances in Neural Information Processing Systems (NeurIPS)},
  year      = {2021}
}

@article{ojala2002multiresolution,
  title   = {Multiresolution gray-scale and rotation invariant texture classification with local binary patterns},
  author  = {Ojala, Timo and Pietik{\"a}inen, Matti and M{\"a}enp{\"a}{\"a}, Topi},
  journal = {IEEE Transactions on Pattern Analysis and Machine Intelligence},
  volume  = {24},
  number  = {7},
  pages   = {971--987},
  year    = {2002}
}

@article{liang2020channel,
  title   = {Channel Compression: Rethinking Information Redundancy among Channels in CNN Architecture},
  author  = {Liang, Jinhua and Zhang, Tao and Feng, Guoqing},
  journal = {arXiv preprint arXiv:2007.01696},
  year    = {2020}
}

@inproceedings{hu2018senet,
  title     = {Squeeze-and-Excitation Networks},
  author    = {Hu, Jie and Shen, Li and Sun, Gang},
  booktitle = {Proc.\ IEEE Conf.\ Computer Vision and Pattern Recognition (CVPR)},
  pages     = {7132--7141},
  year      = {2018}
}

@inproceedings{afchar2018mesonet,
  title={MesoNet: a compact facial video forgery detection network},
  author={Afchar, Darius and Nozick, Vincent and Yamagishi, Junichi and Echizen, Isao},
  booktitle={2018 IEEE International Workshop on Information Forensics and Security (WIFS)},
  pages={1--7},
  year={2018},
  organization={IEEE}
}

@inproceedings{he2016deep,
  title={Deep residual learning for image recognition},
  author={He, Kaiming and Zhang, Xiangyu and Ren, Shaoqing and Sun, Jian},
  booktitle={Proceedings of the IEEE conference on computer vision and pattern recognition (CVPR)},
  pages={770--778},
  year={2016}
}

@inproceedings{tan2019efficientnet,
  title={EfficientNet: Rethinking model scaling for convolutional neural networks},
  author={Tan, Mingxing and Le, Quoc},
  booktitle={Proceedings of the International Conference on Machine Learning (ICML)},
  pages={6105--6114},
  year={2019},
  organization={PMLR}
}

@inproceedings{chollet2017xception,
  title={Xception: Deep learning with depthwise separable convolutions},
  author={Chollet, Fran{\c{c}}ois},
  booktitle={Proceedings of the IEEE conference on computer vision and pattern recognition (CVPR)},
  pages={1251--1258},
  year={2017}
}

@inproceedings{dolhansky2020deepfake,
  title={The DeepFake Detection Challenge (DFDC) Dataset},
  author={Dolhansky, Brian and Howes, Russell and Pflaum, Ben and Baram, Niv and Ferrer, Cristian Canton},
  booktitle={Proceedings of the IEEE/CVF Conference on Computer Vision and Pattern Recognition (CVPR) Workshops},
  pages={1608--1617},
  year={2020}
}

@article{mallat1989theory,
  title={A theory for multiresolution signal decomposition: the wavelet representation},
  author={Mallat, St{\'e}phane},
  journal={IEEE Transactions on Pattern Analysis and Machine Intelligence},
  volume={11},
  number={7},
  pages={674--693},
  year={1989},
  publisher={IEEE}
}

@misc{google2019dfd,
  title        = {Contributing Data to Deepfake Detection Research},
  author       = {{Google AI Blog}},
  year         = {2019},
  howpublished = {\url{https://ai.googleblog.com/2019/09/contributing-data-to-deepfake-detection.html}},
  note         = {Accessed: 2025-09-10}
}

@misc{li2020faceshifter,
  title         = {FaceShifter: Towards High Fidelity and Occlusion Aware Face Swapping}, 
  author        = {Li, Lingzhi and Bao, Jianmin and Yang, Hao and Chen, Dong and Wen, Fang},
  year          = {2020},
  eprint        = {1912.13457},
  archivePrefix = {arXiv},
  primaryClass  = {cs.CV},
  url           = {https://arxiv.org/abs/1912.13457}
}

@inproceedings{li2018ictuadfv,
  title     = {In {Ictu} Oculi: Exposing AI Created Fake Videos by Detecting Eye Blinking},
  author    = {Li, Yuezun and Chang, Ming-Ching and Lyu, Siwei},
  booktitle = {2018 IEEE International Workshop on Information Forensics and Security (WIFS)},
  pages     = {1--7},
  year      = {2018},
  organization = {IEEE},
  doi       = {10.1109/WIFS.2018.8630787}
}

@inproceedings{li2020celeb,
  title     = {Celeb-DF: A Large-scale Challenging Dataset for DeepFake Forensics},
  author    = {Li, Yuezun and Yang, Xin and Sun, Pu and Qi, Honggang and Lyu, Siwei},
  booktitle = {Proceedings of the IEEE/CVF Conference on Computer Vision and Pattern Recognition (CVPR)},
  pages     = {3207--3216},
  year      = {2020}
}

\end{document}